\documentclass{article}
\PassOptionsToPackage{numbers,sort&compress}{natbib}
\usepackage[preprint]{neurips_2025}
\usepackage[utf8]{inputenc}
\usepackage[T1]{fontenc}
\usepackage{hyperref}
\usepackage{url}
\usepackage{booktabs}
\usepackage{amsfonts}
\usepackage{nicefrac}
\usepackage{microtype}
\usepackage{graphicx}
\usepackage{xcolor}
\title{Are Stated Reasoning Steps Causally Load-Bearing?\\
Activation Patching Says Less Often Than Text Editing Does}
\author{%
  Abhiram Bhupatiraju \\
  The University of Texas at Austin \\
  \texttt{ab82836@my.utexas.edu} \\
  \And
  Rayan Nyaupane \\
   The University of Texas at Austin \\
  \texttt{rayannyaupane@utexas.edu} \\
}
\begin{document}
\maketitle
\begin{abstract}
Chain-of-thought (CoT) monitoring assumes that the reasoning a model writes
reflects the computation that directly produces its answer. Previous
faithfulness metrics have been predominantly \emph{behavioral}, as they simply
edit the reasoning text and observe the resulting answer. However, our methodology aims to measure faithfulness \emph{causally} at the
activation level, specifically on self-generated reasoning. Unlike previous
causal audits, which measure degradation, our interventions carry a known
predicted target. In this way, each patch should switch the answer to a
specific counterfactual entity derivable by construction. Specifically, we use synthetic multi-hop lookup tasks (2--6 hops). We patch
the residual stream at the token span where the model states each intermediate
step with the corresponding activations from a counterfactual run. For
Qwen3-4B, $76.9\% \pm 2.8\%$ of stated steps are causally load-bearing (CLB)
at the most responsive mid-network layer (random-position null: $11.3\%$;
patching the underlying prompt fact: $83\%$, so stated steps carry
approximately $96\%$ of the achievable effect).
Moreover, the standard behavioral test on the same items yields $88.2\%$,
which overstates causal faithfulness by $11.4$ percentage points
(item-matched; $111{:}14$ discordant pairs, $p<10^{-15}$) and, for the easiest
items, by up to $20$ percentage points. This gap also has a clear capability dimension. Qwen3-1.7B is far less
causally faithful overall ($54.8\%$), with its faithfulness collapsing as
reasoning depth increases ($68\%$ at 2 hops to $30\%$ at 6), while Qwen3-4B
remains relatively flat. Although stated reasoning can be causally meaningful,
standard behavioral tests tend to overestimate its causal faithfulness,
particularly on easier examples where model reasoning appears most fluent.
\end{abstract}
\section{Introduction}
One of the main oversight mechanisms proposed for increasingly capable language models is CoT monitoring~\cite{korbak2025monitor}. This is where a model's external reasoning~\cite{wei2022} and thus an overseer can audit it. This mechanism, however, is only as good as the link between its stated reasoning and the computation behind the actual answer. For example, a model writes ``Step 2: Grunkum'' and then produces an answer computed through a more internal pathway that never consults the preceding step. Thus, the CoT serves as post-hoc narration rather than pure computation.
Faithfulness has mostly been measured behavioral, where the model paraphrases the reasoning text and observes the changes in the answer~\cite{turpin2023,lanham2023}. However, such behavioral tests are both informative but indirect. For example, when an edited step changes the answer, the model may simply be re-reading its (now inconsistent) transcript. On the other hand, when the model doesn't change its answer, the information may still have been routed internally. In previous works, there have been audits in CoT at the activation level~\cite{bypassing2026,sae2025}. However, these audits measure performance degradation under patching relative to controls. Thus, degradation establishes that the reasoning tokens are significant. Degradation cannot establish that a specific stated step carries the specific content into the answer.
In this work, we aim to close this targeted interchange intervention on self-generated reasoning. Our synthetic task construction gives every intervention a predicted outcome. We aim to build a counterfactual variant for each reasoning step of each item in which that step resolves to a different entity, with the implied final answer derivable in-context by construction. We will then patch the counterfactual step's activations into the clean run at the exact tokens where the model stated the step. We then ask whether the answer flips to the predicted target. In specific. Flip-to-target is a strict criterion that separates ``this step's stated content is casually used downstream'' from ``patching perturbed the computation.''
\textbf{Contributions.}
(1)~A causal faithfulness metric---the causally load-bearing (CLB) rate---based on targeted interchange with known counterfactual targets, with positive, sanity, and specificity controls.
(2)~An item-matched comparison against the standard behavioral text-editing test, quantifying how much the field's prevailing metric overstates causal faithfulness.
(3)~A two-model comparison showing causal faithfulness improves and stabilizes with model capability, while the weaker model's stated reasoning degrades into narration as task difficulty grows.
\section{Related work}
Behavioral faithfulness tests manipulate the inputs or reasoning text. They often bias features that models may not verbalize~\cite{turpin2023}, truncate or corrupt chain-of-thought reasoning~\cite{lanham2023}, and edit self-generated intermediate reasoning structures at scale~\cite{performative2026}. More causal and mechanistic analyses of chain-of-thought include feature-level patching with sparse autoencoders~\cite{sae2025}, as well as layerwise patching of the hidden states associated with CoT tokens, where effects are measured by degradation relative to matched controls~\cite{bypassing2026,breaking2026}. Other work patches CoT hidden states into direct-answer runs to measure how much information those states contain, or probes the model for a latent commitment to an answer before the reasoning process has finished~\cite{hiddenstates2026}.
Our work differs in two main ways. First, our interventions are constructed so that they have a known target answer. This allows us to use a flip-to-target criterion rather than simply measuring degradation in performance. Second, we apply both the causal test and the standard behavioral faithfulness test to the same items, allowing us to directly measure and calibrate the gap between the two. Our approach builds on prior activation-patching methodology~\cite{meng2022,zhang2024patching}.
\section{Method}
\textbf{Task.}
Each item in the task is a multi-hop lookup over an in-context knowledge base, noted as KB, of a pseudo-word entity. The facts are in the form of `The $r$ of $X$ is $Y$,'' with a query that gives a start entity and a relation sequence $r_1, \ldots, r_h$, where $h$ is between 2 and 6, with 50 items each alongside distractor facts. These pseudo-words help eliminate pretraining leakages along with an additional filter that helps to keep every entity to 3 tokens or less. The prompt also further instructs the model to state each intermediate in the form of `Step $k$: '' and finish with ``Answer: .''
Furthermore, for every hop $k$ of every time, we also construct a counterfactual. This counterfactual is a single fact resolving hop $k$ which is further altered so that the preceding step will yield entity $E_k'$. These bridge facts included in the clean KB help to ensure the counterfactual-implied final answer is derivable in-context. We then use a validator to re-derive the gold and all counterfactual answers from said rendered text, which was able to achieve $1000/1000$ passing.
\textbf{Intervention.}
A clean run is defined as having both a correct answer and the exact format. Within each of these runs, there were $224/250$ usable runs (for Qwen3-4B~\cite{qwen3}) and $120/250$ ( for Qwen3-1.7B). For each hop $k$:
\begin{enumerate}
\item First, generate a countercall run and locate the exact token span where the model states $E_k'$
\item Then, we cache the residual-stream activations (specifically the block outputs) at the specific span
\item Next, we rerun the clean prompt generation up until the end of the clean step-$k$ span while continuously substituting the cached activations at layer $L$. We do this with the spans end-aligned and greedily continuing
\item Finally, we score the result as a CLB if it equals the counterfactual-simpled answer. However, we disrupt this answer if it changes ss. Otherwise, we keep it unchanged if it remains equal to the desired, golden answer.
\end{enumerate}
Specifically, all generations are greedy bf16 with fixed seeds. Additionally, a random-position null runs alongside each intervention, which uses the same activations patched at a random non-step span.
\textbf{Behavioral baseline.}
Within the same items and hops, we replaced $E_k$ with $E_k'$ in the generated text. This is then truncated after the edited step and thus continued with the generation. This result is scored identically
\textbf{Controls.}
\begin{enumerate}
\item Prompt-fact patch (positive): Patch the KB fact's object-entity activations from the counterfactual prompt into the clean prompt. After regenerating the full CoT, this acts as an upper bound on achievable flips.
\item Post-answer patch (sanity): Patch after the answer is emitted. This must yield 0 flips.
\item Shuffled-target (specificity): Patch a donor entity from a different item's run that is absent from the current KB. Such flips to the donor should be rare.
\end{enumerate}
\section{Results}
\textbf{Stated steps are largely load-bearing.} For Qwen3-4B, patching the
stated-step span changes the final answer to the specific predicted
counterfactual in 78.3\% $\pm$ 2.8\% of 851 interventions at layer 4 and
76.9\% $\pm$ 2.8\% at layer 8, our main layer, compared with 8.9-12.5\% for
random-position controls. The positive control, which patches the
underlying knowledge-base fact directly, flips 83\% of cases, so stated-step
patching reaches about 96\% of this ceiling. Step representations therefore
hold nearly the full causal content of the facts they summarize. The
post-answer sanity control produces 0\% flips at every layer, and
shuffled-target patches flip to the donor entity in only 1-15\% of cases.
Disruption, meaning the answer changed but not to the target, stays at or
below 19\% for step patches throughout.
\begin{figure}[t]
  \centering
  \includegraphics[width=\linewidth]{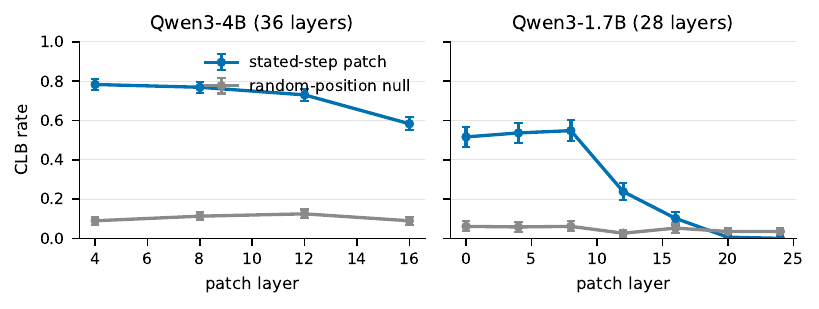}
  \caption{CLB rate by patch layer (95\% CIs), plotted against the
  random-position null. The effect peaks in early and middle layers and
  disappears by layer 20 of 36 (4B) and by layers 12-16 of 28 (1.7B). Once
  the continuation has consumed the stated step, later-layer content no
  longer changes the outcome.}
  \label{fig:layers}
\end{figure}
This effect is concentrated in a specific band of the network. It is
confined to early and middle layers and disappears by layer 20 of 36 in the
4B model and by layers 12-16 of 28 in the 1.7B model (earlier layers
approach a token-identity swap, discussed in Limitations). This pattern fits
a picture where the continuation reads stated-step content from
shallow-to-mid representations at the step position, after which the answer
is assembled from what was already read and later-layer content no longer
matters.
\textbf{The behavioral test overstates this faithfulness.} On the same
items, text editing flips the answer in 88.2\% $\pm$ 2.1\% of 873 cases.
Matched item-by-item at layer 8, the behavioral score exceeds the causal
score by 11.4 points, with 111 items flipping under the behavioral test
alone against 14 flipping under the causal test alone (McNemar test,
$p < 10^{-15}$). Behavioral flips are close to a superset of causal flips.
\begin{figure}[t]
  \centering
  \includegraphics[width=\linewidth]{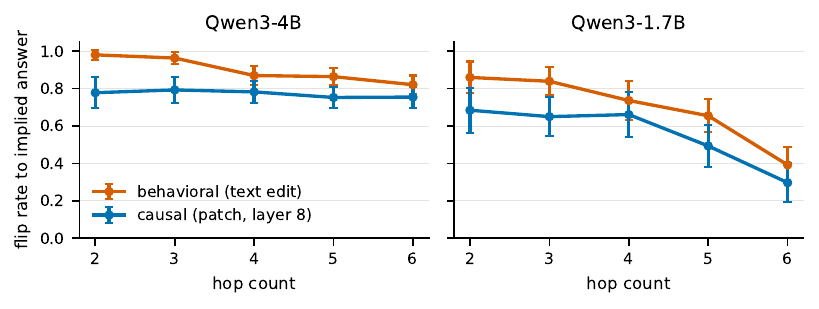}
  \caption{Flip rate to the predicted counterfactual answer by hop count
  (95\% CIs). The behavioral text-editing test (orange) is above the causal
  CLB rate (blue) at every hop count, with the largest gap on easy items.
  Qwen3-4B (left) stays flat across difficulty; Qwen3-1.7B's causal
  faithfulness (right) drops with depth while its behavioral score remains
  much higher.}
  \label{fig:hops}
\end{figure}
The size of this gap is not constant. It is largest where the model
performs best, 20.2 points at 2 hops, narrowing to 6.6 points at 6 hops:
the behavioral test overstates faithfulness most in exactly the regime
where fluent reasoning would otherwise inspire the most confidence.
\textbf{Causal faithfulness tracks model capability rather than problem
difficulty, except when the model is weak.} Qwen3-4B's CLB rate stays
within a narrow band (75-79\%) across 2 to 6 hops. Qwen3-1.7B is both less
causally faithful overall (54.8\% $\pm$ 5.3\% at its best middle layer,
against a behavioral score of 67.6\%, a 9.9-point gap) and highly sensitive
to difficulty: its CLB rate falls from 68.4\% at 2 hops to 29.7\% at 6 hops,
while its behavioral score stays 9 to 19 points higher across the same
range. The claim that stated steps become decorative as problems get harder
holds for the weaker model and fails for the stronger one on this task
family.
\section{Discussion and limitations}
Roughly three quarters of the stated steps are causally load-bearing by a strict flip-to-target criterion, which is close to the method's own ceiling. The main warning lies in the measurement, where the behavioral editing test that dominates the current practice overstates causal faithfulness on every condition we ran. This occurred the most on the easiest items, specifically the regime where the fluent CoT inspires the most trust. Such pipelines where monitoring and evaluations rely on behavioral faithfulness scores should treat them as upper bounds. On the other hand, the capability effect cuts the other way as causal faithfulness rose with model scale in our pair (though two models showcase a trend rather than a law).
\textbf{Limitations.}
Our domain lies in synthetic single-relation lookup. Naturalistic reasoning, such as math and code, may differ. We also only study one model family, Qwen3~\cite{qwen3}, specifically at small scales using greedy decoding. Early-layer patches partially overlap with the token-identity replacement. This is why we headline the mid-band layer. Though, a fuller treatment would patch attention and MLP outputs separately. Furthermore, span alignment truncates when clean and counterfactual entity spans differ in length. Behavioral and causal interventions, while item-matched, differ in what the model sees. For example, text edits are visible in-context while activation paths are not.


\begin{thebibliography}{12}\small
\bibitem{korbak2025monitor} T. Korbak et al.
\newblock Chain of thought monitorability: A new and fragile opportunity for AI safety.
\newblock arXiv:2507.11473, 2025.
\bibitem{turpin2023} M. Turpin, J. Michael, E. Perez, S. Bowman.
\newblock Language models don't always say what they think.
\newblock NeurIPS, 2023.
\bibitem{lanham2023} T. Lanham et al.
\newblock Measuring faithfulness in chain-of-thought reasoning.
\newblock arXiv:2307.13702, 2023.
\bibitem{bypassing2026} A. Sathyanarayanan, A. Nagarsekar, A. Rathore.
\newblock Bypassing the rationale: Causal auditing of implicit reasoning in language models.
\newblock arXiv:2602.03994, 2026.
\bibitem{performative2026} W. Li, F. Yang, A. Hazarika, S. A. Mehta, K. Onoue.
\newblock When reasoning traces become performative: Step-level evidence that chain-of-thought is an imperfect oversight channel.
\newblock arXiv:2605.11746, 2026.
\bibitem{breaking2026} O. Somov et al.
\newblock Breaking the chain: A causal analysis of LLM faithfulness to intermediate structures.
\newblock arXiv:2603.16475, 2026.
\bibitem{hiddenstates2026} H. Mehrafarin, A. Parekh, I. Konstas.
\newblock When chain-of-thought fails, the solution hides in the hidden states.
\newblock arXiv:2604.23351, 2026.
\bibitem{sae2025} X. Chen, A. Plaat, N. van Stein.
\newblock How does chain of thought think? Mechanistic interpretability of chain-of-thought reasoning with sparse autoencoding.
\newblock arXiv:2507.22928, 2025.
\bibitem{meng2022} K. Meng, D. Bau, A. Andonian, Y. Belinkov.
\newblock Locating and editing factual associations in GPT.
\newblock NeurIPS, 2022.
\bibitem{zhang2024patching} F. Zhang, N. Nanda.
\newblock Towards best practices of activation patching in language models.
\newblock ICLR, 2024.
\bibitem{qwen3} Qwen Team.
\newblock Qwen3 technical report.
\newblock arXiv:2505.09388, 2025.
\bibitem{wei2022} J. Wei et al.
\newblock Chain-of-thought prompting elicits reasoning in large language models.
\newblock NeurIPS, 2022.
\end{thebibliography}
\end{document}